\documentclass{article} 
\usepackage{nips15submit_e,times}
\usepackage{amsmath, amssymb, amsthm, xspace, subfigure,algorithmic,algorithm, dsfont, hyperref, url, graphicx,widetext}
\usepackage[numbers]{natbib}

\newcommand{\figref}[1]{Fig.~\ref{fig:#1}}
\newcommand{\tblref}[1]{Table~\ref{tbl:#1}}

\def\real{\mathbb{R}}
\newcommand{\squeeze}{\vspace{-2mm}}

\theoremstyle{definition}
\newtheorem{defn}{Definition}[section]

\title{Quantification in-the-wild: data-sets and baselines}

\author{
Oscar Beijbom \hspace{10mm} Judy Hoffman \hspace{10mm} Evan Yao \hspace{10mm} Trevor Darrell \\ 
University of California, Berkeley, California\\
\texttt{\{obeijbom, jhoffman, trevor\}@eecs.berkeley.edu} 
\hspace{3mm}
\texttt{\{evanyao\}@berkeley.edu}
\vspace{-3mm}
\And 
Alberto Rodriguez - Ramirez \hspace{10mm} Manuel González - Rivero \hspace{10mm} Ove Hoegh - Guldberg \\
Global Change Institute, The University of Queensland, Australia \\
\texttt{\{alberto.rodriguez, m.gonzalezrivero, oveh\}@uq.edu.au} \vspace{-3mm}
}

\nipsfinalcopy 

\begin{document}

\maketitle
\vspace{-3mm}
\begin{abstract}
Quantification is the task of estimating the class-distribution of a data-set. While typically considered as a parameter estimation problem with strict assumptions on the data-set shift, we consider quantification in-the-wild, on two large scale data-sets from marine ecology: a survey of Caribbean coral reefs, and a plankton time series from Martha's Vineyard Coastal Observatory. We investigate several quantification methods from the literature and indicate opportunities for future work. In particular, we show that a deep neural network can be fine-tuned on a very limited amount of data (25 - 100 samples) to outperform alternative methods.
\end{abstract}
\squeeze
\squeeze
\section{Introduction}
\squeeze
The term `quantification' was introduced by Forman~\cite{forman2008quantifying} and defined as the task of estimating the class-distribution of a categorical data-set. In a typical scenario, a classifier is trained on a set of labeled `source' samples, and applied to a new set of `target' samples. This is challenging in the presence of `data-set shifts'\cite{moreno2012unifying} where the underlying probability function of the source data differs from that of the target data. Contrary to classification, which is studied both in the presence and absence of domain shift, quantification is only of interest under data-set shift, since otherwise, the class-distribution can be estimated directly from the labeled source data.

Quantification is important for two principled reasons. First the class-distribution itself, rather than the full labeling, is the desired end-product in many applications. This occurs for example in sampling surveys, where repeated quantification is required to assess spatial or temporal patterns~\cite{forman2008quantifying, sampling}. Second, in applications where a full labeling is required, the class-distribution of the target data can be used to re-calibrate a classifier trained on the source data~\cite{royer2015classifier, saerens2002adjusting}.

Quantification has been studied under the class-distribution shift assumption~\cite{saerens2002adjusting, forman2008quantifying, du2014semi}, meaning that the source and target class conditional distributions are the same~\cite{moreno2012unifying}. As we shall see, this is a rather strong assumption and may not hold in practice. In another line of work, domain adaptation (DA) has been studied under the assumption that the source and target class-conditional distributions are `similar'~\cite{jing_literature}. In this work, the class-distribution shift is often controlled for by experimenting on class-balanced data-sets~\cite{saenko_adapting}.

We introduce two large-scale data-sets from marine ecology, where quantification arise naturally and where automation is imperative for ecological analysis. We evaluate efficacy of several quantification methods from the literature.

\squeeze
\subsection{Problem statement}
\squeeze
Let $x\in \mathbb{R}^d$ be d-dimensional input samples, and $y \in {1, \hdots, c}$ class labels. We assume that a large number of samples and classes, $\{(x_i, y_i)\}_{i=1}^n$, are available in a source domain defined by some joint probability function $p(x, y)$. We also assume that a large number of unlabeled samples, $\{x_i\}_{i=1}^{n'}$, are available in a target domain, defined by different probability function $p'(x, y) \neq p(x, y)$. The general goal of quantification is to estimate the probability distribution over classes in the target domain: $p'(y) \equiv q \in \real^c$. However, the problem statement as defined above is intractable if (a) the domain shift is arbitrary and (b) there are no labeled samples in the target domain. It is therefore typically studied under one of two relaxations:

\begin{defn}
\textbf{Unsupervised Quantification:} In unsupervised quantification, the data-set shift is assumed to be a pure class-distribution shift, i.e. $p(y) \neq p'(y)$, but $p(x|y) = p'(x|y)$. Alternatively, the data-set shift is assumed to be `small', and the unlabeled set of target samples, $\{x_i\}_{i=1}^{n'}$, is used to align the internal feature representation of a machine learning algorithm.
\end{defn}

\begin{defn}
\textbf{Supervised Quantification:} In supervised quantification, no explicit assumptions are made on the data-set shift, but it is assumed that a small amount of samples are available in the target domain, $\{(x_i, y_i)\}_{i=1}^b \in p'(x, y)$. 
\end{defn}
For supervised quantification, we only consider methods where the labeled target samples are selected randomly, leaving the design of active sampling methods to future work.
\squeeze
\subsection{Related work}
\squeeze
For unsupervised quantification, a straight-forward method is classify \& count~\cite{forman2008quantifying} where a classifier, $f$ is trained using the source data and then used to estimate $\hat{q}_c = \frac{1}{n'} \sum_{i = 1}^{n'} \mathds{1} (f(x_i), c)$. The classifier, $f$ can also be adapted using unsupervised adaption methods~\cite{Tzeng_ICCV2015,hal_domain}. Further, unsupervised quantification has been studied extensively under class-distribution shift~\cite{moreno2012unifying,forman2008quantifying, saerens2002adjusting}. This work follow one of two main strategies. The first, introduced by Saerens et al.~\cite{saerens2002adjusting}, and refined by~\cite{du2014semi}, derive an EM algorithm for maximizing the likelihood of the target data, $p'(x)$ by iterative updates of $\hat{p}'(y)$ and $\hat{p}'(y|x)$. The second, discussed by several authors~\cite{forman2008quantifying, saerens2002adjusting, solow2001estimating, beijbom2014cost}, rely on the miss-classification rates (confusion matrix) estimated on the source data to adjust the estimated counts on the target data. Forman extended this method with several heuristic demonstrating significant performance increase for binary quantification~\cite{forman2008quantifying}.

For supervised quantification, simple random sampling~\cite{sampling} can be utilized to achieve an unbiased estimate the class-distribution: $\hat{q}_c = \frac{1}{b} \sum_{i = 1}^b \mathds{1} (y_{s_i}, c)$ where $s$ is a vector of randomly permuted indices and $b$ the annotation budget. Simple random sampling doesn't utilize the classifier, $f$, but it can be incorporated using auxiliary sampling designs, through offset~\cite{beijbom2014cost} or ratio~\cite{royall1981empirical} estimators. In these methods, some property (e.g. bias) of the classifier is estimated from the labeled subset, and then used to adjust the prediction on the whole target set. Other methods include adapting the classifier to operate in the target domain, typically achieved by modifying the internal feature representation of the classifier~\cite{Tzeng_ICCV2015}.

\squeeze
\section{Data-sets}
\squeeze
We introduce two large-scale, image data-sets from marine ecology. In both, repeated sets of collected survey images require quantification, and manual annotation is unfeasible due to the vast amounts of collected data. In both, a large set of labeled source images are available to train a classifier, and several randomly selected smaller sets are available for evaluation. Each such set is denoted a test `cell' and the goal is to achieve accuracy quantification across all test cells. Domain shifts occur naturally in these data-sets, so that the class appearance, $p(x|y)$ and class-distributions, $p(y)$, varies across the test cells. However, the extent of these variations differ between the two data-sets as discussed below. A data-set overview is given in \tblref{data-sets}, and all data can be downloaded from \url{www.eecs.berkeley.edu/~obeijbom/quantification_dataset.html}.
\squeeze
\subsection{Plankton population survey}
\squeeze
\textbf{Background:} The Imaging Flow Cytobot (IFCB), is an \emph{in-situ} instrument for measuring plankton populations~\cite{olson2007submersible, ifcb_cvpr}. The IFCB is installed at an offshore tower at 4 m below water level at the Martha's Vineyard Coastal Observatory, and collects images by automatically drawing seawater from the environment. From this stream of images ($\sim100$k day$^{-1}$) the ecologists need to quantify the daily plankton class-distribution. Manual annotation of the complete data stream is unfeasible, and is currently restrained to two randomly chosen hours each month. While insufficient for a complete ecological analysis, these randomly selected, fully annotated, image-sets are ideal for evaluation of quantification methods.

\textbf{Details:} We formalize the Plankton Survey quantification benchmark as follows. All IFCB labeled data from 2006-2013 is considered as pertaining to the source domain. The 21 randomly selected hours of annotated data from 2014 are the test cells. Only classes with $>1000$ total samples are included in this benchmark, leaving around 3.3 million total labeled samples across $33$ classes. The Plankton Survey data-set is dominated by a class-distribution shift (Figs. \ref{fig:ifcb_class_counts}, \ref{fig:ifcb_appearance_shift}).
\squeeze
\subsection{Coral reef survey}
\squeeze
\textbf{Background:} The XL Catlin Seaview Survey (XL CSS) is an ambitious project to monitor the world's coral reefs~\cite{gonzalez2014catlin}. Using underwater scooters, 2 kilometer of reef-scape is imaged each dive, with approximately one image meter$^{-1}$. The XL CSS has surveyed the Great Barrier Reef, the Coral Sea, the Caribbean, the Coral Triangle, the Maldives, and  Chagos, and captured over 1 million photographs. From this images set, ecologists are interested in quantifying the percent cover of key benthic substrates for each set of $30$ consecutive images~\cite{gonzalez2014catlin}. Percent cover for each image is estimated by classifying 50 patches extracted at random row \& column locations in each image as pertaining to one of $32$ classes~\cite{pante2012getting}, for a total of $\sim 1500$ patches across the $30$ images. Similarly to the Plankton Survey, we can think of these sets of patches as cells, each requiring quantification. Manual quantification of all cells is unfeasible: the images from the Caribbean alone would require 30 person-years to annotate~\cite{gonzalez2014catlin}. 

\textbf{Details:} We formalize the Coral Survey quantification benchmark as follows. A training-set of $324732$ annotated patches extracted from $1505$ images constitute the source domain. In addition, $15$ randomly selected sets of $30$ consecutive images, each with $50$ annotated patches, are the test cells (\figref{map}). Both class appearance and class-distribution varies across the test cells (since they are drawn from different location across the Caribbean), meaning that the data-set shifts are more complex than for the IFCB data-set (Figs. \ref{fig:css_class_counts}, \ref{fig:css_appearance_shift}). 

\squeeze
\begin{table}[t]
\centering
\caption{Data-set summary. n = number, avg. = average}
\small
\begin{tabular}{| l | c | c | c | c |}
\hline 
  Data-set & n train samples & n test cells & avg. n samples cell$^{-1}$ & n classes \\
   \hline \hline
  Plankton Survey & 3.3m & 21 & 14248 & 33 \\
   \hline
  Coral Survey & 325k & 15 & 1480 & 32 \\
   \hline
\end{tabular}
\label{tbl:data-sets}
\end{table}

\squeeze
\subsection{Performance evaluation}
\squeeze
Let $q^{(m)} = P(y)$ be the normalized ground-truth class-distribution: $q^{(m)} \in \real^c, \sum_{j=1}^c q^{(m)}_j = 1 ~ \forall m$ for sampling cell $m$, and $\hat{q}$ the estimated distribution. We measure the distance between $q^{(m)}$ and $\hat{q}^{(m)}$ using the Bray-Curtis distance, which is commonly used in ecology. For normalized class counts, it reduces to the l1-norm:
$h^{\mathrm{BC}}(q^{(m)}, \hat{q}^{(m)}) = \frac{|q^{(m)}-\hat{q}^{(m)}|_1}{2}$. The utility of a quantification method is evaluated by the average Bray-Curtis distance across the sampling cells.
\squeeze
\section{Experiments}
\squeeze
For all experiments, we train a Convolutional Neural Network, $f$, of the Alexnet network architecture~\cite{krizhevsky2012imagenet}, on the source data using Caffe~\cite{jia2014caffe}. More details are given in~\cite{ifcb_cvpr, beijbom2014cost}.

\begin{figure}[htb] 
\squeeze
\begin{center}
\includegraphics[width=.35 \linewidth]{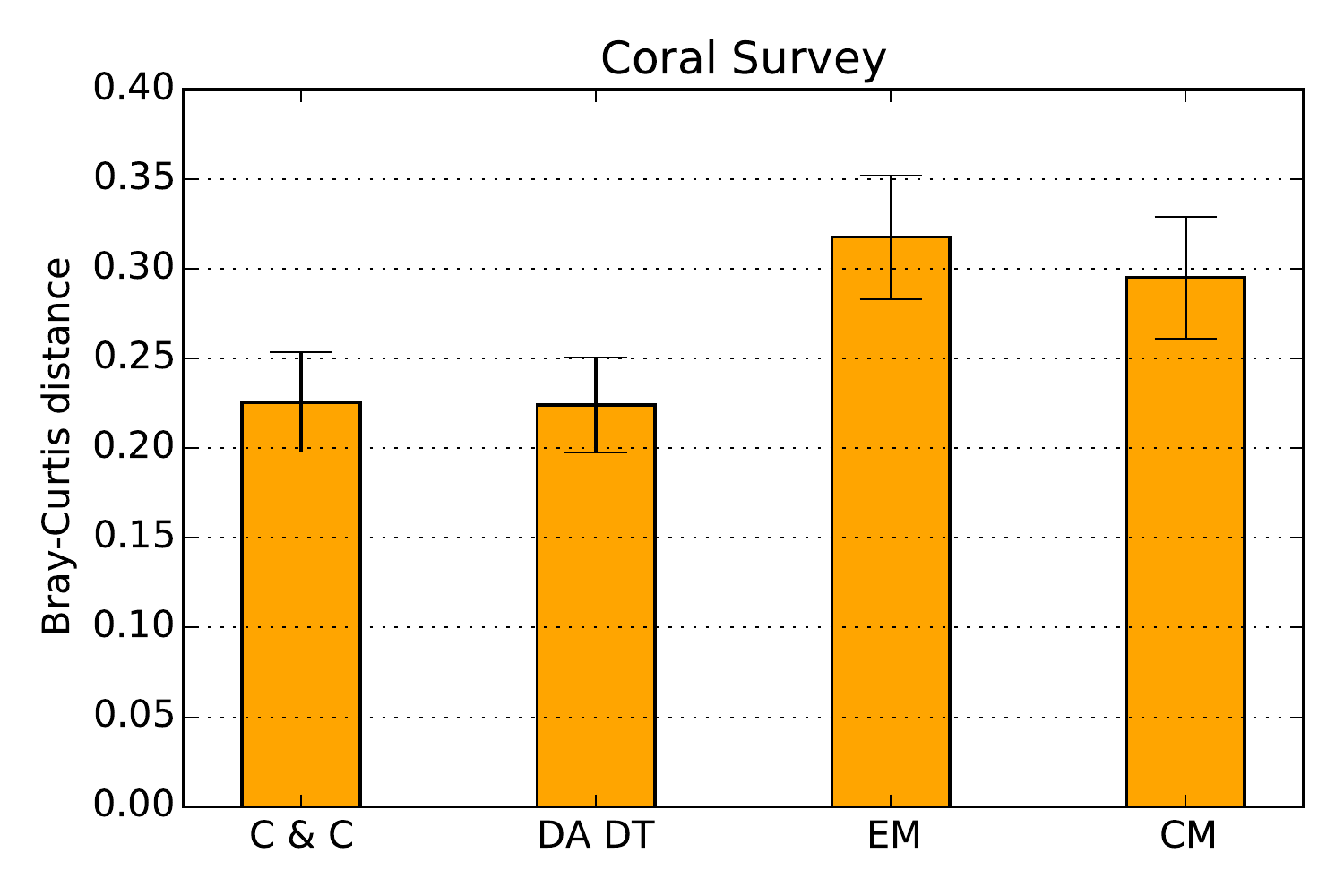}
\hspace{10mm}
\includegraphics[width=.35 \linewidth]{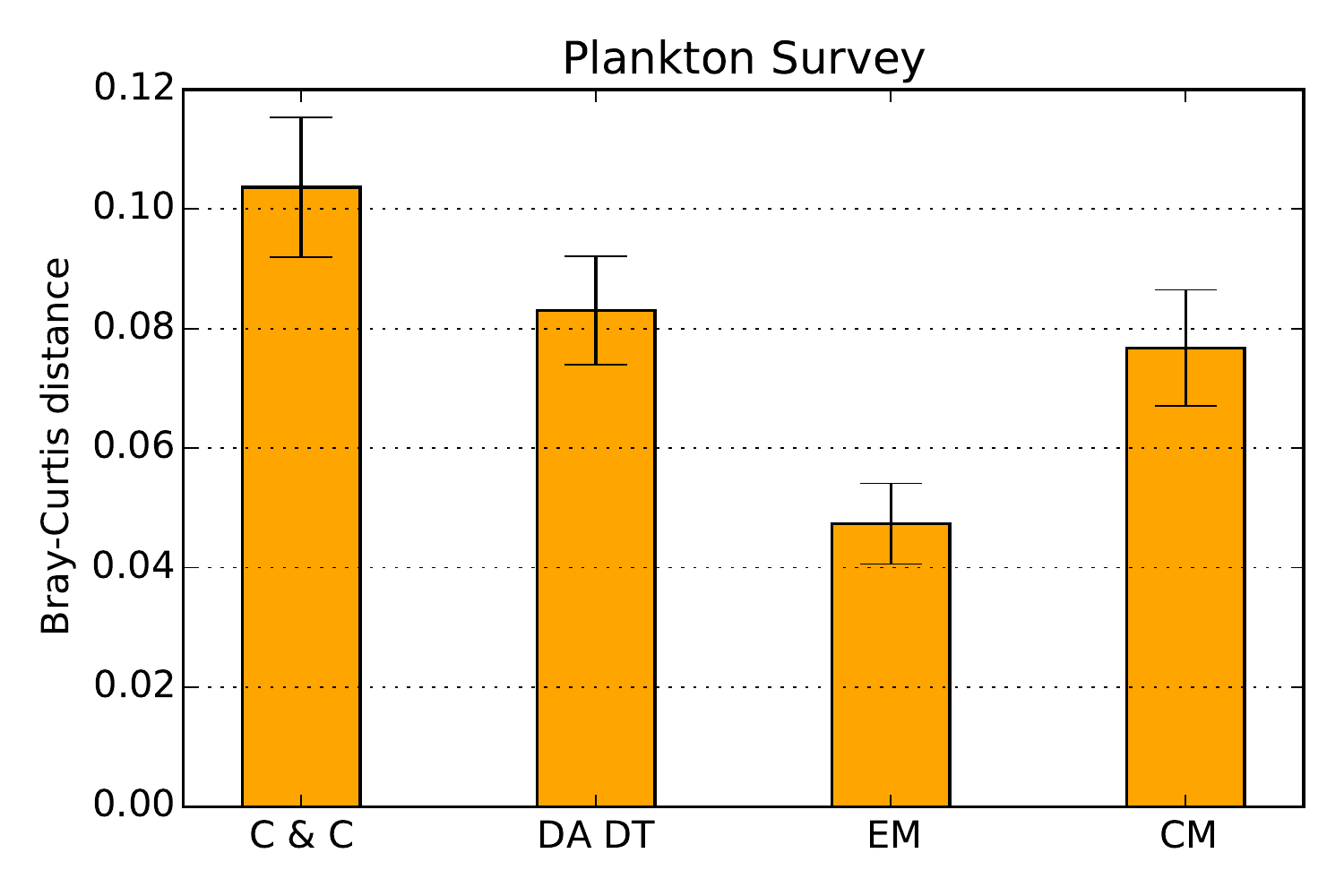}
\end{center}\vspace{-4mm}
\caption{\textbf{Unsupervised results.} Quantification errors displayed as mean $\pm$ SE for Classify \& Count~\cite{forman2008quantifying}, the unsupervised Deep Transfer DA method of~\cite{Tzeng_ICCV2015}, distribution matching using the EM algorithm~\cite{saerens2002adjusting}, and for correction using Confusion Matrix inversion~\cite{forman2008quantifying, solow2001estimating, saerens2002adjusting}.}
\label{fig:res-unsuper}
\squeeze
\end{figure}

\textbf{Unsupervised quantification:} We evaluate four unsupervised quantification methods. Applying Classify \& Count is straight-forward, and creates a natural baseline (\figref{res-unsuper}). The EM-algorithm of~\cite{saerens2002adjusting} is also evaluated (\figref{em_convergence}) along with the confusion-matrix (CM) correction method of~\cite{forman2008quantifying, solow2001estimating, saerens2002adjusting}. The latter method requires inverting the confusion-matrix, which can be problematic for multi-class problems, since inversion requires full rank. We therefore apply the abundance correction for each class, $i$ independently, by mapping the CM to a binary $2 \times 2$ matrix before inverting and estimating $\hat{q}_i$. We then normalized $\hat{q}$ so that $\sum_{j=1}^c \hat{q}_j = 1$. Finally, we use a recent unsupervised domain-adaptation method which adapts the source net to the unlabeled data from each cell~\cite{Tzeng_ICCV2015}.

\textbf{Supervised quantification:} We investigate quantification performance for budgets, $10 < b < 150$ samples, which well covers the feasible range of what is economical for the respective surveys. Random sampling is used to establish an upper bound on the error, and the offset estimator is included as a simple improvement~\cite{beijbom2014cost}. We also experimented with a ratio estimator~\cite{sampling}, but this is impractical for small sample sizes~\cite{beijbom2014cost}. Further, we used a DA baseline ('DA mix'). In this method, the classifier, $f$ was further fine-tuned on a mixture of $75\%$ source and $25\%$ target data (drawn from the $b$ samples) for $\sim3$ epochs. Finally, the supervised DA method of~\cite{Tzeng_ICCV2015} was evaluated.

\begin{figure}[htb] 
\begin{center}
\includegraphics[width=.45 \linewidth]{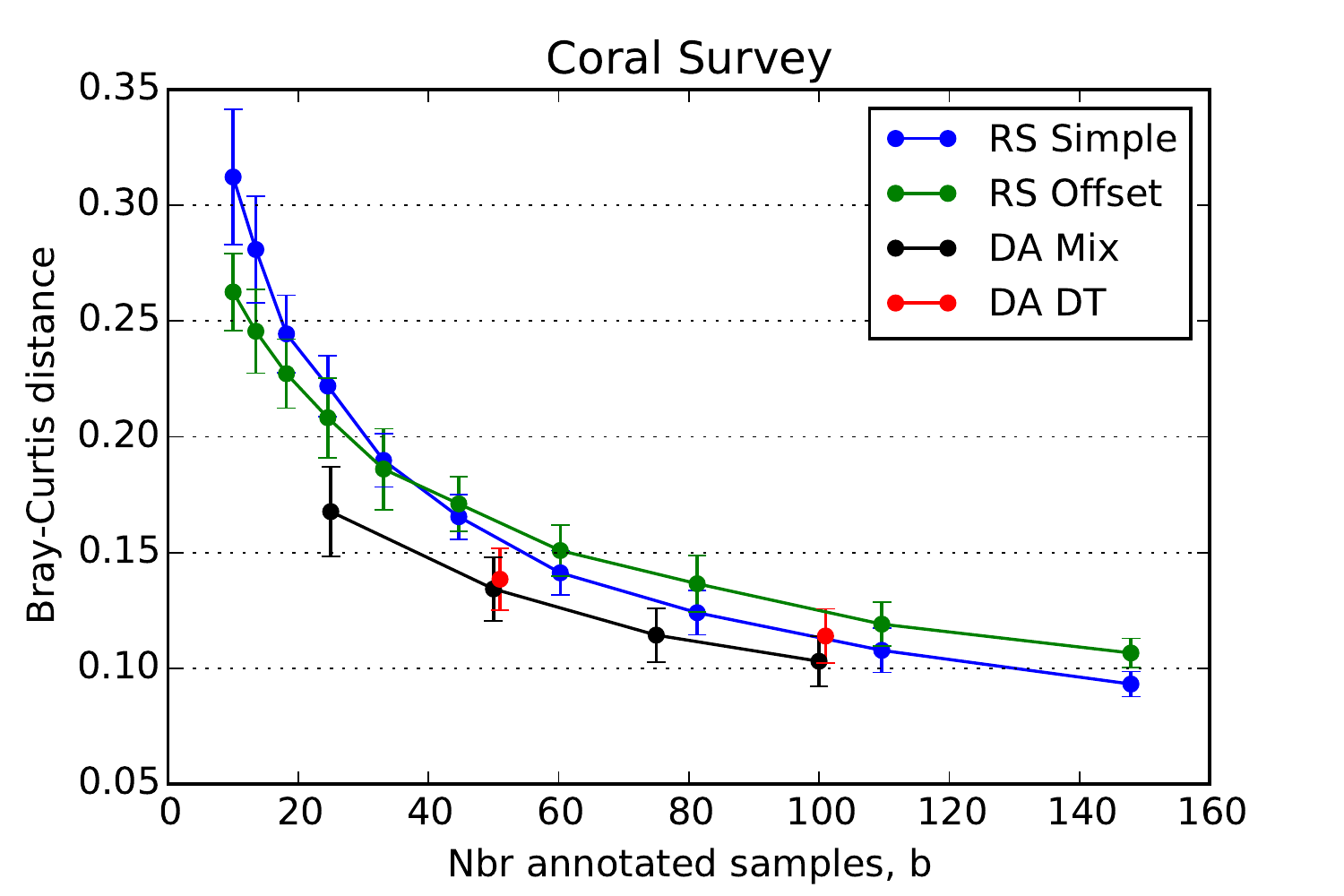}
\hspace{4mm}
\includegraphics[width=.45 \linewidth]{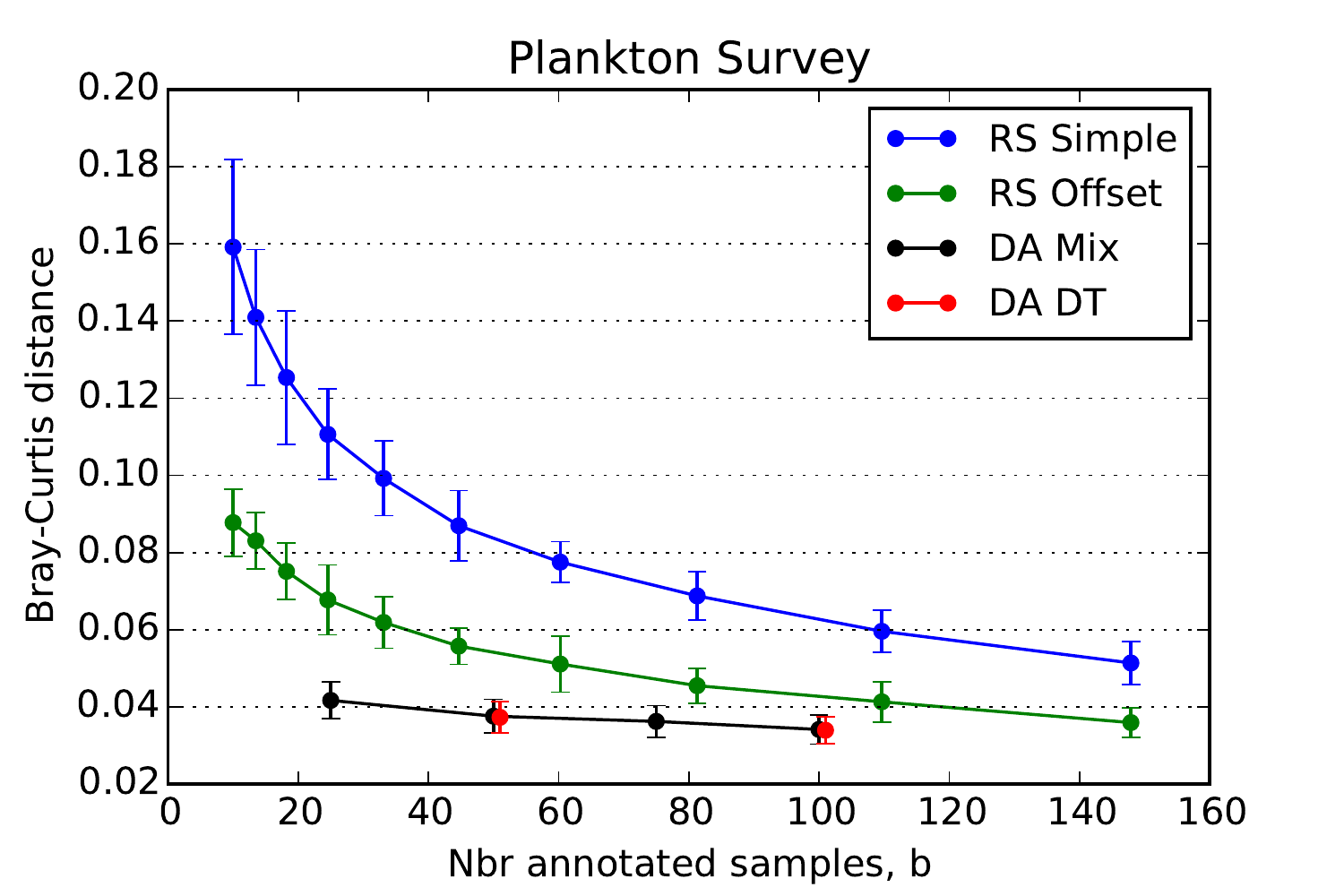}
\end{center}\vspace{-4mm}
\caption{\textbf{Supervised results.} Quantification errors displayed as mean $\pm$ SE for Simple random sampling~\cite{sampling}, Offset sampling~\cite{beijbom2014cost}, DA mix, and Deep Transfer DA ~\cite{Tzeng_ICCV2015}}
\label{fig:res-super}
\squeeze
\end{figure}

\textbf{Results:} For unsupervised quantification, our results indicate that the appropriate method depends on the nature of the data-set shift. For the Plankton Survey, the EM and CM correction methods work well, significantly lowering the estimation errors (\figref{res-unsuper}). The EM method~\cite{saerens2002adjusting} outperformed the CM method~\cite{forman2008quantifying}, suggesting that this approach is more appropriate for high number of classes. However, for the Coral Survey, where the class-distribution shift assumption is violated, the CM and EM corrections corrupt the results, and simply counting the raw classification is preferable. The Deep Transfer DA method~\cite{Tzeng_ICCV2015} was able to capture the data-set shift for the Plankton Survey, but produced inferior quantification compared to the CM and EM methods. Also note that the quantification results are, in general, stronger for the Plankton Survey, since it is an easier classification task with smaller data-set shift.

Fur supervised quantification, the two DA methods outperformed the random sampling baselines, in particular for smaller annotation budgets (\figref{res-super}). The adaptation method of~\cite{Tzeng_ICCV2015} performed on par with DA mix. Among the random sampling methods, the offset estimator clearly outperformed simple random sampling on the Plankton Survey, but performed on-par for the Coral Survey. This is expected, as the hybrid estimator performs better when the classification errors are small, which they are in the Plankton Survey (\figref{res-super};~\cite{beijbom2014cost}).

\textbf{Discussion:} In pure class-distribution shift situations, as with the Plankton Surveys, the EM algorithm of~\cite{saerens2002adjusting} worked well, achieving a mean Bray-Curtis distance of $4.7 \pm 3.2\%$. Achieving such accurate quantification through simple random sampling would require $b \approx 150$ samples. The DA mix method, which achieved $3.7 \pm 0.5\%$ at $b = 50$ and $4.1 \pm 0.4\%$ at $b = 25$, makes better use of the supervision. It is a compelling alternative overall since it performed well also on the more challenging Coral Survey data. This is important since, in a real-world situation, one may not now \emph{a-priori} what type of data-set shift to expect for the new target data.

The fact that fine-tuning of a deep neural network can be achieved with such small amount of target data ($25$ samples) is surprising, and deserves further investigation. Further, while the EM method presented here didn't perform very strongly, Forman suggested several improvements for binary classification~\cite{forman2008quantifying}. We were unable to generalize these to the multi-class case, but it deserves attention. Finally, we think active sampling methods offer much promise, with collected samples utilized either to correct the raw classification counts, or for fine-tuning model parameters.

\textbf{Acknowledgments:} This work was supported by the National Oceanic and Atmospheric Administration grant No. NA10OAR4320156 and by the XL and Catlin Group Limited, Global Change Institute. We gratefully acknowledge the support of NVIDIA for their hardware donations. 

\bibliographystyle{plain}
\bibliography{active_quantification}

\pagebreak
\begin{widetext}
\begin{center}
\textbf{\Large Supplementary Information: Quantification in-the-wild: data-sets and baselines}
\end{center}
\setcounter{equation}{0}
\setcounter{figure}{0}
\setcounter{table}{0}
\setcounter{page}{1}
\makeatletter
\renewcommand{\theequation}{S\arabic{equation}}
\renewcommand{\thefigure}{S\arabic{figure}}
\renewcommand{\bibnumfmt}[1]{[#1]}
\renewcommand{\citenumfont}[1]{#1}
\appendix
\section{Supplementary data-set information}
\subsection{Coral Survey}
Image from the Catlin Seaview Survey can be accessed from \url{catlinseaviewsurvey.com} and \url{http://globalreefrecord.org/}. The particular sites studied in this paper are indicated in~\figref{map}. From each image, 30 patches are cropped and classified. Some samples patches are shown in \figref{coral_images.pdf}. The data-set shift in the Coral Survey data-set is shown in \figref{css_appearance_shift} and \figref{css_class_counts}.

\begin{figure}[ht] 
\begin{center}
\includegraphics[width=.70 \linewidth]{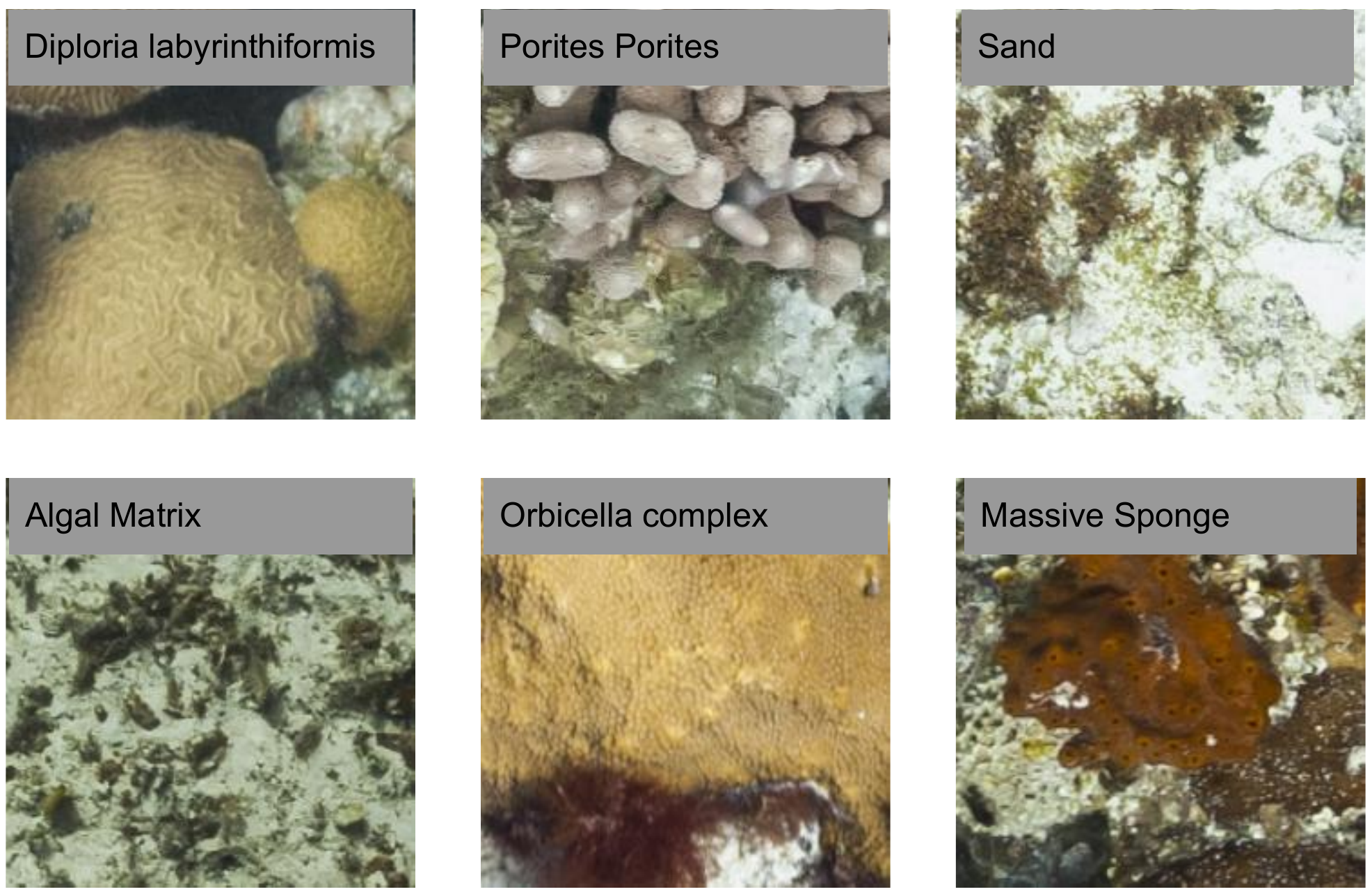}
\end{center}
\caption{Sample images from the Coral Survey}
\label{fig:coral_images.pdf}
\end{figure}

\begin{figure}[ht] 
\begin{center}
\includegraphics[width=.70 \linewidth]{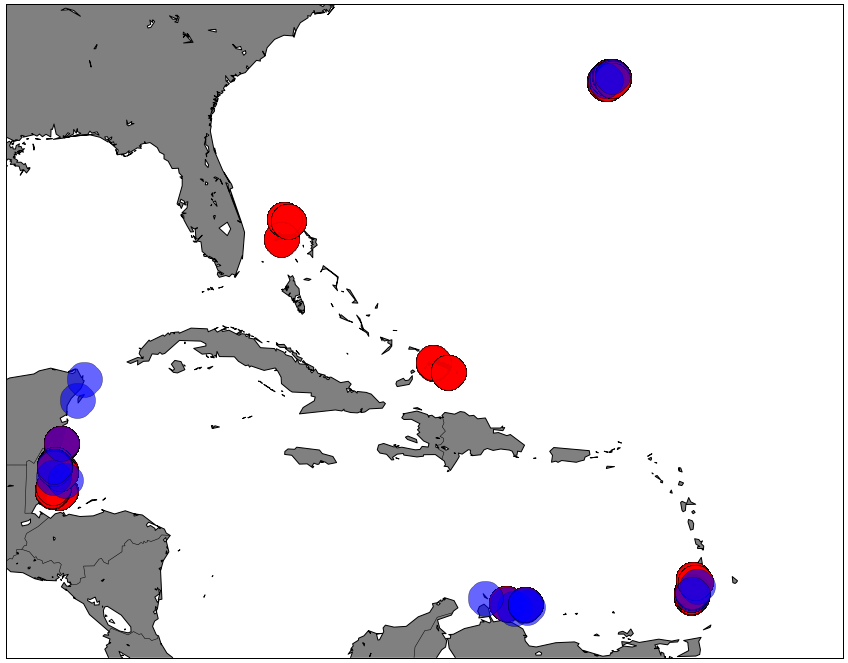}
\end{center}
\caption{Caribbean locations used as for the Coral Survey benchmark. Blue circles indicate location of the test-cells and red circles the training images.}
\label{fig:map}
\end{figure}

\begin{figure}[ht] 
\begin{center}
\includegraphics[width=.90 \linewidth]{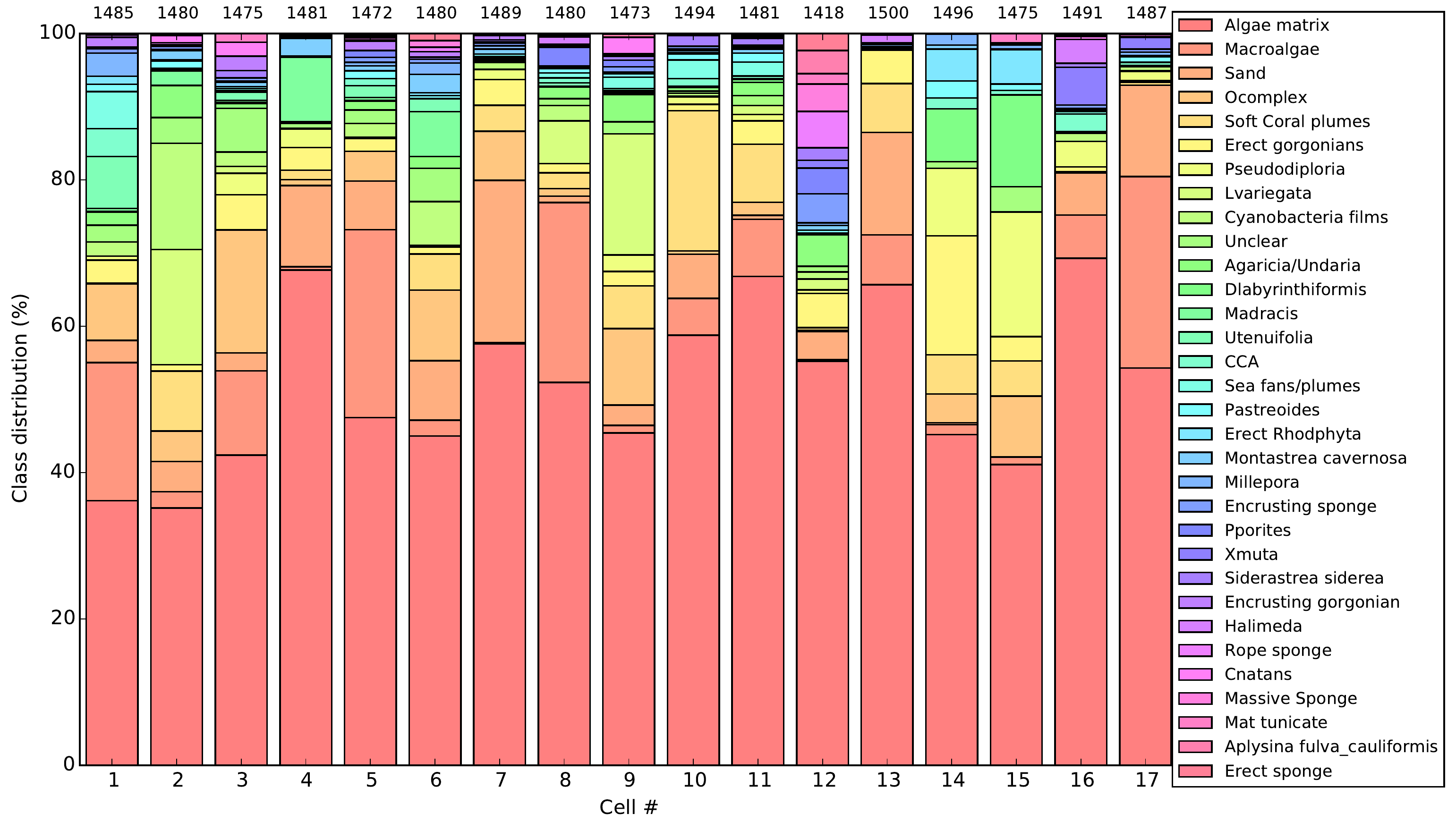}
\end{center}
\caption{Class-distribution for the Coral Survey test cells. Total number of cell members is indicated above each cell}
\label{fig:css_class_counts}
\end{figure}

\begin{figure}[ht] 
\begin{center}
\includegraphics[width=.90 \linewidth]{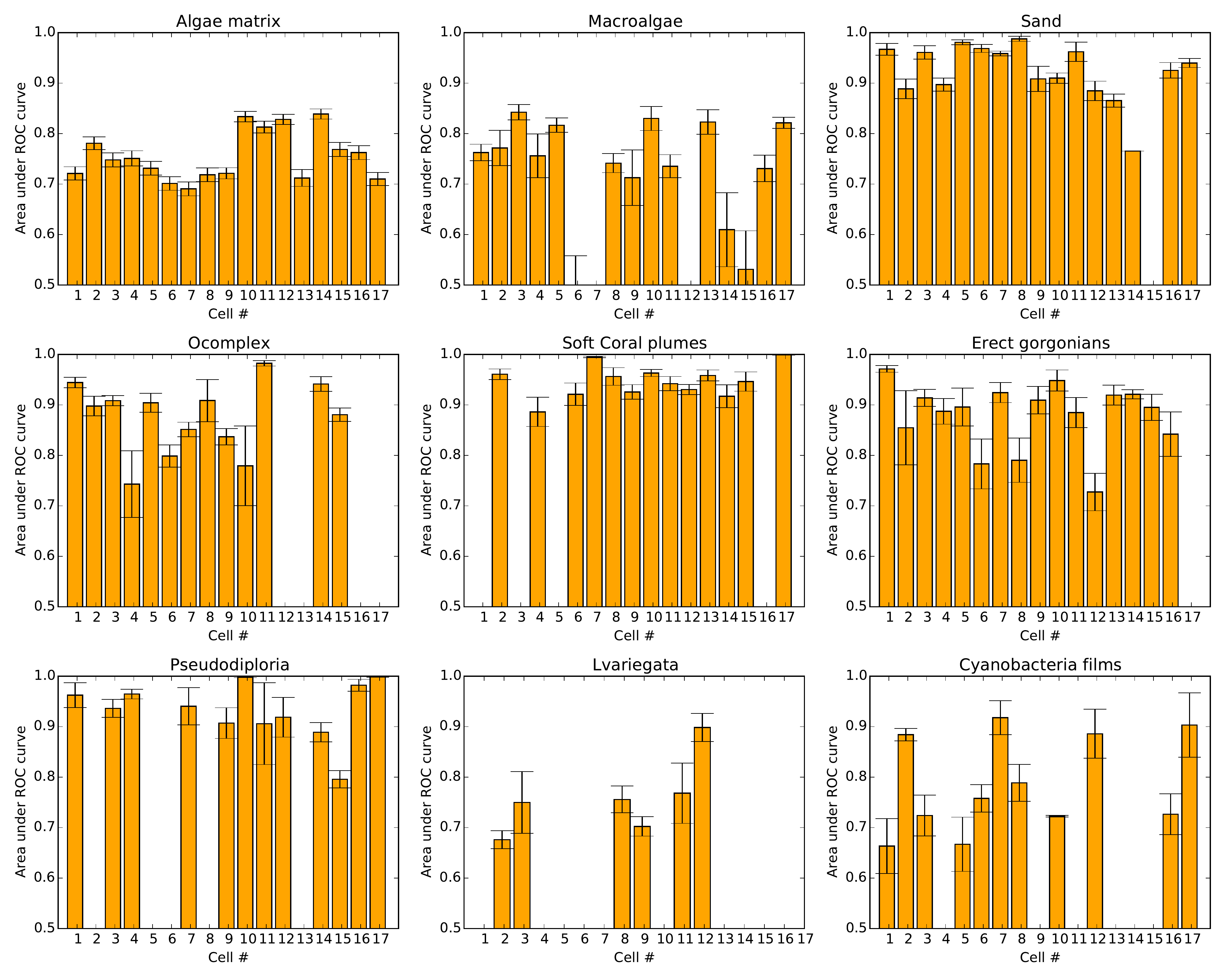}
\end{center}
\caption{Area under ROC curve (mean $\pm$ SD) for the 9 most abundance classes in the Coral Survey. The large difference indicate a significant shift in appearance, $p(x|y)$, between the cells.}
\label{fig:css_appearance_shift}
\end{figure}
\clearpage

\subsection{Plankton Survey}
The IFCB data-set can be downloaded at \url{https://github.com/hsosik/WHOI-Plankton} from which an interactive image viewer can be reached. Some sample images are shown in \figref{plankton_images2.pdf}. 

\begin{figure}[ht] 
\begin{center}
\includegraphics[width=.80 \linewidth]{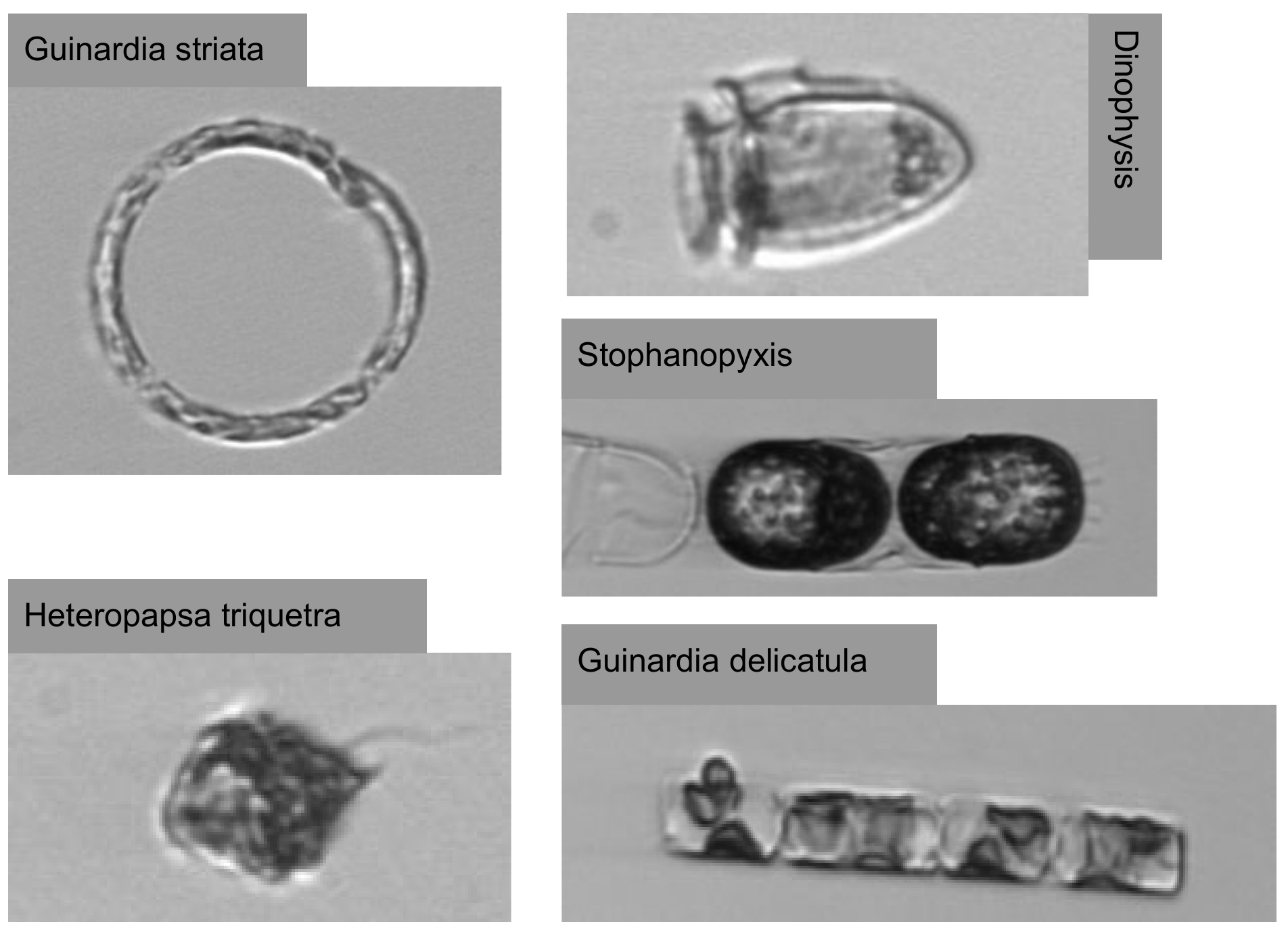}
\end{center}
\caption{Sample images from the Plankton Survey}
\label{fig:plankton_images2.pdf}
\end{figure}

\begin{figure}[ht] 
\begin{center}
\includegraphics[width=.80 \linewidth]{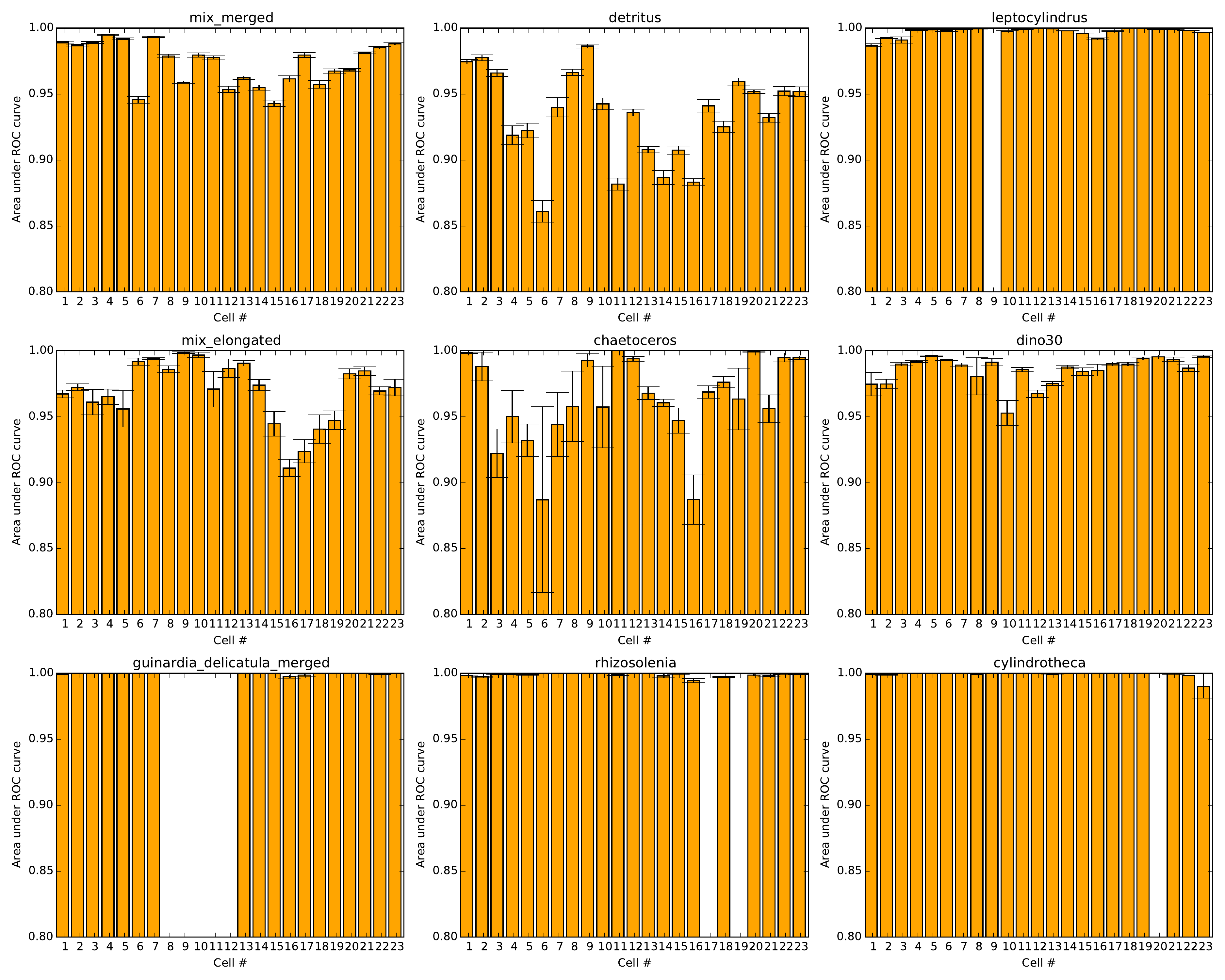}
\end{center}
\caption{Area under ROC curve (mean $\pm$ SD) for the 9 most abundance classes in the Plankton Survey. Note that the smaller y-axis range compared to \figref{css_appearance_shift}.}
\label{fig:ifcb_appearance_shift}
\end{figure}

\begin{figure}[ht] 
\begin{center}
\includegraphics[width=1 \linewidth]{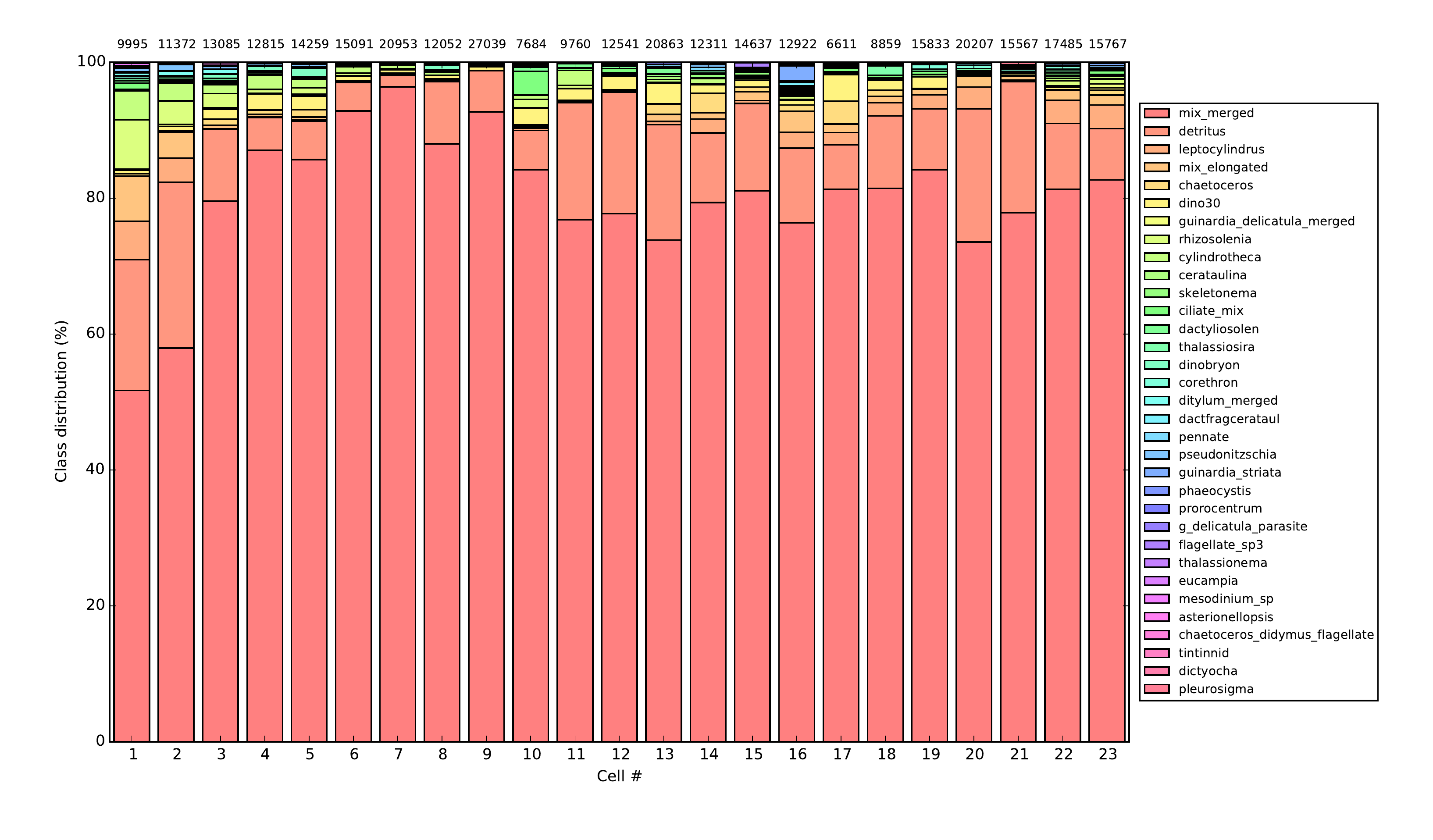}
\includegraphics[width=1 \linewidth]{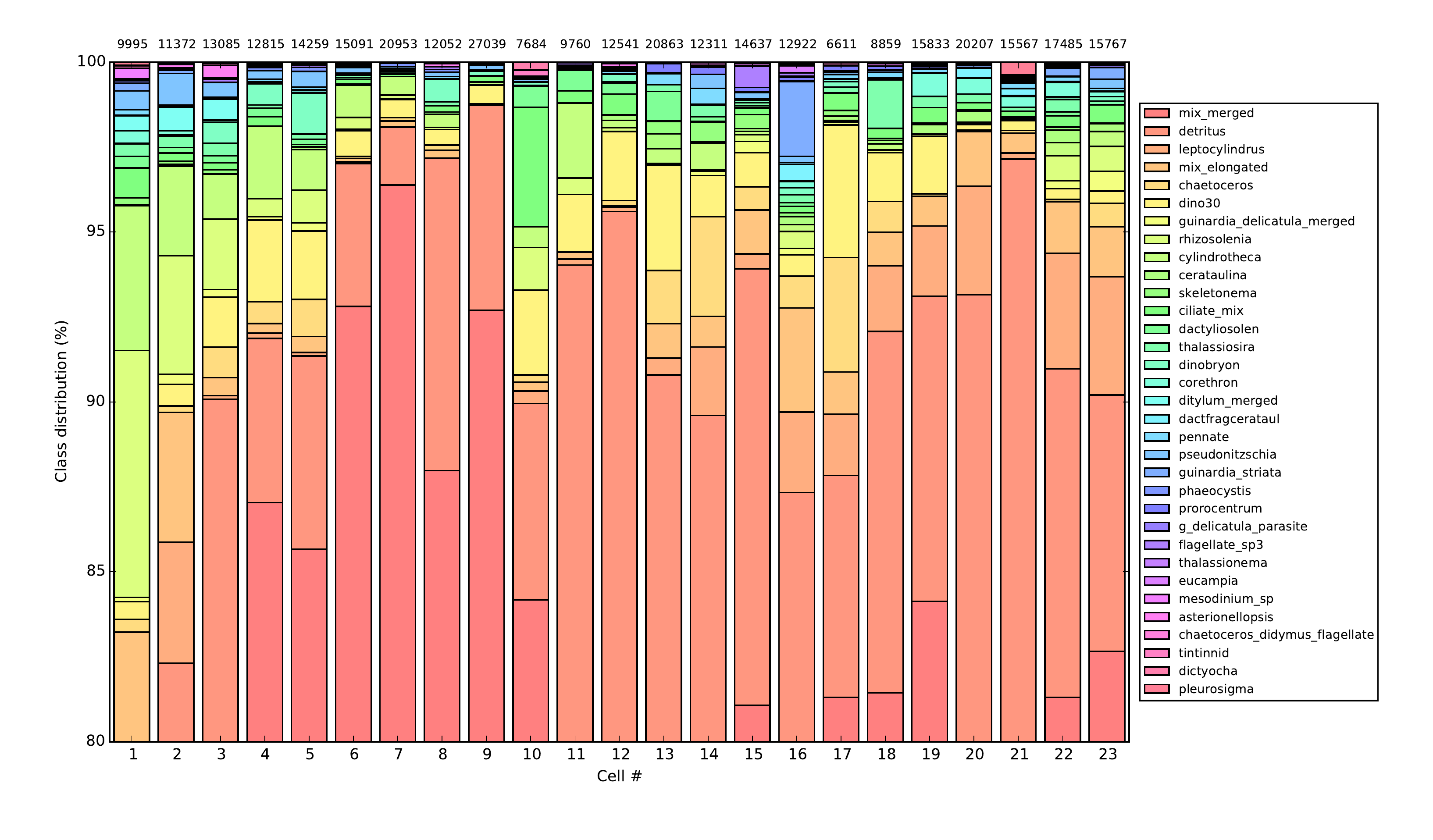}
\end{center}
\caption{Class-distribution for the Plankton Survey test cells, sorted by overall class abundance. Total number of cell members is indicated above each cell. Top: full distribution illustrating the dominance of the `mix' class, bottom: highlight of the range between $80\%$ and $100\%$}
\label{fig:ifcb_class_counts}
\end{figure}

\clearpage
\section{Supplementary results}
We include a convergence plot for the EM algorithm of~\cite{saerens2002adjusting} in~\figref{em_convergence}
\begin{figure}[ht] 
\begin{center}
\includegraphics[width=.95 \linewidth]{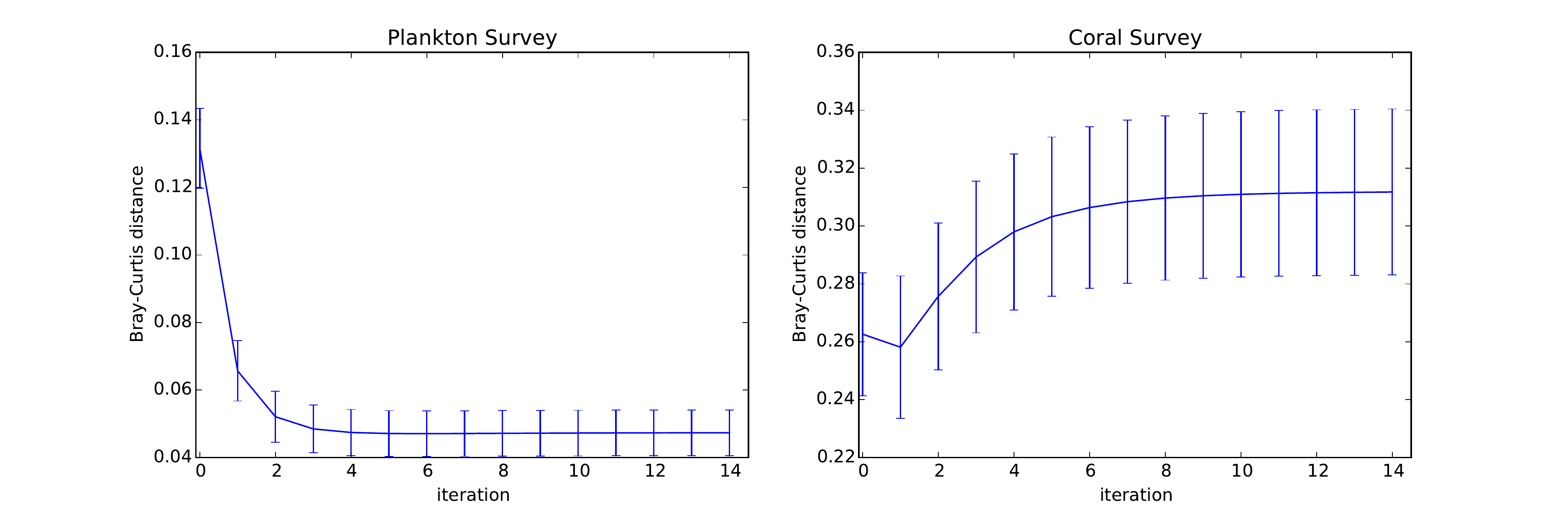}
\end{center}
\caption{Convergence of the EM algorithm~\cite{saerens2002adjusting}. Since the class-distribution shift assumption of this method is not satisfied for the Coral Survey, it converges to a unsatisfactory state.}
\label{fig:em_convergence}
\end{figure}

\end{widetext}
\bibliographystyle{plain}
\bibliography{active_quantification}

\end{document}